\documentclass[letterpaper, 10 pt, conference]{ieeeconf}
\IEEEoverridecommandlockouts
\usepackage{amsmath}
\usepackage{amssymb}
\usepackage{balance}
\usepackage{booktabs}
\usepackage{graphicx}
\usepackage{siunitx}
\usepackage{textcomp}
\usepackage[absolute,showboxes]{textpos}
\usepackage{xcolor}
\usepackage[caption=false, font=footnotesize, subrefformat=parens, labelformat=parens]{subfig}
\graphicspath{{./img/}}
\DeclareSIUnit\vehicle{veh}
\DeclareSIUnit\lane{lane}
\TPMargin*{3pt}

\newcommand\copyrighttext{
	\footnotesize
	\noindent
	\textcopyright\,2022 IEEE.
	Personal use of this material is permitted.
	Permission from IEEE must be obtained for all other uses, in any current or future media, including reprinting/republishing this material for advertising or promotional purposes, creating new collective works, for resale or redistribution to servers or lists, or reuse of any copyrighted component of this work in other works.}%
\newcommand\copyrightnotice{%
	\begin{textblock*}{7in}(0.75in,0.25in)
		\copyrighttext
	\end{textblock*}
}

\title{\LARGE \bf
An Enhanced Graph Representation for Machine Learning Based Automatic Intersection Management
}

\author{
	Marvin~Klimke$^{1,2}$,
	Jasper~Gerigk$^{1}$,
	Benjamin~V\"olz$^{1}$, and
	Michael~Buchholz$^{2}$%
\thanks{$^{1}$The authors are with the Robert Bosch GmbH, Corporate Research, D-71272 Renningen, Germany. E-Mail: {\tt\small\{marvin.klimke, benjamin.voelz\}@de.bosch.com}}%
\thanks{$^{2}$The authors are with the Institute of Measurement, Control and Microtechnology, Ulm University, D-89081 Ulm, Germany. E-Mail: {\tt\small michael.buchholz@uni-ulm.de}}%
\thanks{Part of this work was financially supported by the Federal Ministry for Economic Affairs and Climate Action of Germany within the program "Highly and Fully Automated Driving in Demanding Driving Situations" (project LUKAS, grant numbers 19A20004A and 19A20004F).}%
}

\begin{document}

\maketitle
\copyrightnotice
\thispagestyle{empty}
\pagestyle{empty}

\begin{abstract}
The improvement of traffic efficiency at urban intersections receives strong research interest in the field of automated intersection management.
So far, mostly non-learning algorithms like reservation or optimization-based ones were proposed to solve the underlying multi-agent planning problem.
At the same time, automated driving functions for a single ego vehicle are increasingly implemented using machine learning methods.
In this work, we build upon a previously presented graph-based scene representation and graph neural network to approach the problem using reinforcement learning.
The scene representation is improved in key aspects by using edge features in addition to the existing node features for the vehicles.
This leads to an increased representation quality that is leveraged by an updated network architecture.
The paper provides an in-depth evaluation of the proposed method against baselines that are commonly used in automatic intersection management.
Compared to a traditional signalized intersection and an enhanced first-in-first-out scheme, a significant reduction of traversal duration is observed at varying traffic densities.
Finally, the generalization capability of the graph-based representation is evaluated by testing the policy on intersection layouts not seen during training.
The model generalizes virtually without restrictions to smaller intersection layouts and within certain limits to larger ones.
\end{abstract}

\section{Introduction}
\label{sec:intro}

Connected automated driving has the potential to significantly improve traffic efficiency and safety on highways and in urban areas.
By employing a wireless communication link, connected vehicles (CVs) and connected automated vehicles (CAVs) announce their presence and can share perception data.
In urban areas, providing edge computing resources becomes viable, opening up the opportunity of maintaining a local environment model on an edge server.
This allows a fleet of CVs within the operational area to be aware of each other, including cases where vehicle-bound sensor systems are limited by occlusion effects, which are highly prevalent at urban intersections.
Prior research in \cite{buchholz2021handling} shows that an environment model provided by an edge server can be used by a CAV to smoothly merge onto a priority road.
Without this external information, the CAV's planning algorithm typically requires the vehicle to come to a stop before being able to safely merge into a gap.

\begin{figure}
	\centering
	\includegraphics[width=\linewidth]{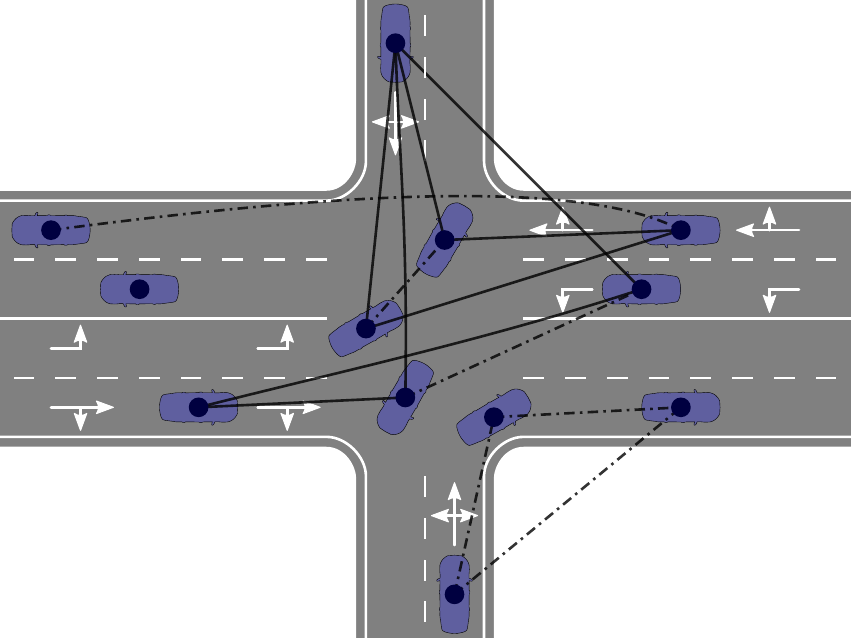}
	\caption{Graph-based scene representation for automatic intersection management. Each vehicle is mapped to a vertex that are connected by two types of edges, depending on the pairwise conflict relations.}
	\label{fig:intro}
\end{figure}

Automatic intersection management (AIM) describes approaches aimed at improving traffic efficiency by collectively controlling multiple vehicles at an intersection.
Most prior research on AIM apply non-learning reservation or optimization-based algorithms.
Machine learning based behavior planning so far concentrated mainly on single ego vehicles.
Only very few consider cooperative multi-agent planning.
In the present work, we leverage recent advances in machine learning and graph neural networks (GNNs) for managing multi-lane intersections, as depicted in Fig.~\ref{fig:intro}.
Building upon the approach presented in \cite{klimke2022cooperative}, we improve the model in key aspects, resulting in the following contributions:
\begin{itemize}
\item Extension of the graph-based scene representation and network architecture by introducing edge features,
\item Providing an in-depth evaluation of our approach against multiple baselines including traffic lights,
\item Demonstrating the model's ability to generalize to intersections not encountered during training.
\end{itemize}

The remainder of the paper is structured as follows:
Section~\ref{sec:sota} discusses the state of the art in AIM.
Our improvements to the graph-based representation and network architecture are given in Section~\ref{sec:approach}, while Section~\ref{sec:sim} introduces the simulation environment used for training and evaluation.
In Section~\ref{sec:eval}, we present our comprehensive evaluation on various intersection layouts and against multiple baselines.
Section~\ref{sec:conclusion} concludes the paper.

\section{Related Work}
\label{sec:sota}

There is a large body of literature on existing approaches to AIM of which we present a selection of commonly used paradigms.
Many works can be allocated to reservation-based systems \cite{nichting2019explicit,nichting2020space,dresner2008multiagent} or optimization schemes \cite{li2018near-optimal,malikopoulos2018decentralized,kamal2015vehicle-intersection}.

A more extensive overview of AIM literature is given in \cite{zhong2020autonomous}, which categorizes the works according to the degree of centralization.
In a fully centralized scheme, a single coordination unit performs multi-agent planning over all vehicles to cross the intersection while acting as the communication partner for each of them.
As a first step towards decentralized AIM, certain vehicles, e.g. platoon leaders, may act as a proxy in the communication with the coordination unit, being responsible for vehicles directly following them.
A fully distributed AIM is characterized by the lack of a central coordination unit.
Instead, the vehicles have to negotiate a cooperative plan on their own.

In \cite{nichting2019explicit}, a distributed negotiation procedure for cooperative lane change maneuvers is presented.
A CAV may request surrounding vehicles to keep a designated area free to safely perform the lane change, if such a maneuver is deemed beneficial.
This approach is extended for usage at intersections \cite{nichting2020space} by triggering a cooperation request when planned paths conflict on the intersection area.
By conducting test drives with two testing vehicles, the feasibility of the proposed approach is demonstrated under low traffic density.

The tile-based reservation system proposed in \cite{dresner2008multiagent} employs a centralized first-in-first-out (FIFO) policy for assigning clearance to the requesting vehicles.
In the evaluation, the authors show a significant benefit in delay per vehicle with increasing traffic density.
Multiple variants of the FIFO~policy are benchmarked against traditional yield or signalized intersections.
Partitioning the available space into tiles that can only be used by a single road user at a time might lead to suboptimal exploitation of the intersection area.

Optimizing the 2D trajectory of a CAV with various constraints leads to a nonlinear programming problem, whose complexity rises with increasing number of vehicles in the scene.
The optimization-based AIM system published in \cite{li2018near-optimal} addresses this issue by defining a set of standard cases that are solved offline.
To apply the precomputed solutions to an arbitrary traffic scene online, a predefined formation is requested before the vehicles enter the intersection area.
The authors acknowledge that solving all standard cases for a multi-lane four-way intersection takes too much time, even when done offline.
Alternative approaches assume the vehicles to follow predefined lanes laterally, while optimizing the longitudinal motion.
The distributed energy-optimizing approach \cite{malikopoulos2018decentralized} further disallows turning maneuvers and driving on two conflicting paths at the same time.
In \cite{kamal2015vehicle-intersection}, the longitudinal control of vehicles is performed by a centralized intersection coordination unit employing a model predictive control scheme.
Both works compare their results with traditional signalized intersections and show a reduction in delay times and fuel consumption.
Apart from reducing the delay, the maximum intersection capacity is increased.
All of these optimization-based approaches face the unfavorable scaling of computational demand with increasing number of road users.

Other approaches to AIM include, for instance, the application of a platooning concept \cite{morales_medina2018cooperative}.
Between a pair of vehicles on conflicting paths, a virtual inter-vehicle distance is calculated and used to control the vehicles' velocity to cross the conflict point with sufficient clearance.
According to the authors, this approach requires significant adaptions to be used in mixed traffic, i.e., simultaneous presence of automated and human-driven vehicles.
In \cite{gradinescu2007adaptive}, it is proposed to equip an adaptive traffic light controller with a communication link to CVs.
By incorporating the detailed information on the traffic demand on different inflow lanes, the actuation of the traffic lights is continuously optimized.
The benefit of this approach is demonstrated by simulating two urban intersection at rush hour, showing a reduction in delay for crossing the intersection.
By employing traffic lights, non-connected road users can be instructed, which allows the application in mixed traffic.

Machine learning experiences high research interest for the application in prediction as well as planning for a single ego vehicle in automated driving \cite{zhu2021survey}.
Reinforcement learning (RL) is used for single ego behavior planning, as demonstrated for urban intersections \cite{capasso2021end--end} or highway lane changes \cite{hart2020graph}.
In the latter work, the authors propose to use a graph-based representation of the semantic environment of the ego vehicle and a fitting GNN for processing.
In cooperative AIM, however, learning-based algorithms have rarely been used.
The application of supervised learning, also known as imitation learning, is limited by the lack of ground truth data for cooperative maneuvers.

In \cite{wu2019dcl-aim}, an RL agent is trained to select the most appropriate action from a discrete action space, which is restricted to collision-free actions by using a tile-based reservation scheme.
Execution of the policy is done decentralized on all agents and thus does not benefit from explicit communication and cooperation between agents.
The authors evaluate this approach against a traditional signalized intersection and a FIFO policy.
In our previous work~\cite{klimke2022cooperative}, we proposed the first centralized approach to AIM leveraging RL and a graph-based scene representation.
The learned planner shows a significant performance benefit in vehicle flow rate and a reduction of induced stops, both in synthetic simulations and based on real-world traffic data.
This paper settles on these results and improves the model to be applied to larger intersection layouts and those that were not seen during training.

\section{Proposed Approach}
\label{sec:approach}

In this section, the main improvements to the machine learning based AIM scheme presented in \cite{klimke2022cooperative} are introduced.
The adaptions of the scene graph, including the addition of edge features, are outlined in Section~\ref{ssec:graph}.
Section~\ref{ssec:network} then describes the changes to the GNN architecture required to leverage the enhanced input representation.

Like in our previous work, we regard the cooperative planning problem as a Markov decision process (MDP) and continue using an RL-based approach.
The MDP is defined as the tuple $(S,\,A,\,T,\,R)$, where $S$ denotes the set of states.
$A = [a_{\mathrm{min}},\,a_{\mathrm{max}}]^N$ describes the action space consisting of a bounded interval of desired acceleration values for each of the $N$ agents currently in the scene.
The transition function $T(s\,,a\,,s')$ denotes the probability of changing from state $s \in S$ to $s' \in S$ when applying action $a \in A$.
The fitness of a chosen action~$a$ executed in state~$s$ is given by the reward function $R(s,\,a)$, which in this work is taken from \cite{klimke2022cooperative}.
Since vehicles may appear and vanish from the scene at any time, the dimensionalities of the state space and the action space vary with the number of currently present agents $N$.
In the current work, the TD3 actor-critic RL algorithm \cite{fujimoto2018addressing} for continuous actions is again used for training the centralized planning policy.

\subsection{Graph-based Representation}
\label{ssec:graph}

\begin{figure}
	\centering
	\includegraphics[width=\linewidth]{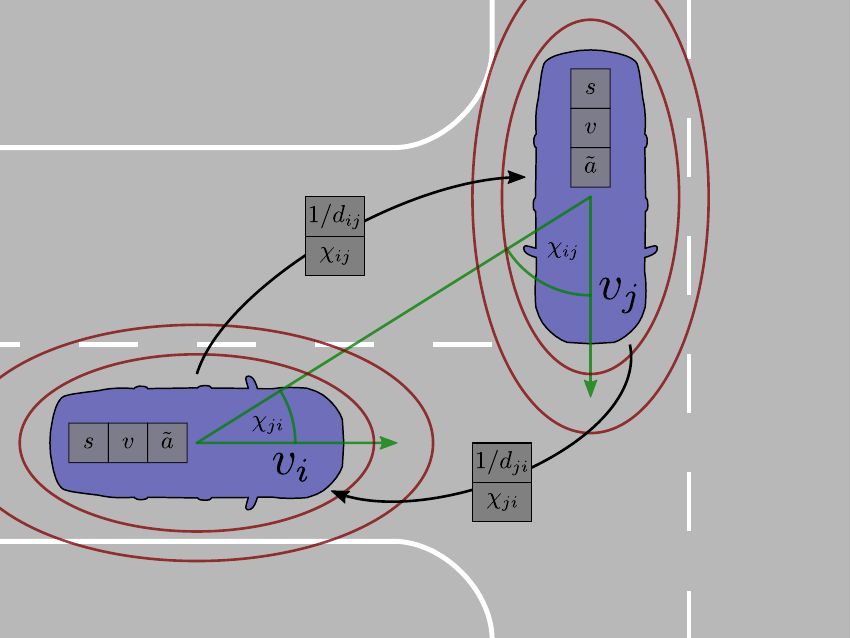}
	\caption{The composition of input vertex and edge features is illustrated by a pair of vehicles. Each vertex feature vector is composed of the longitudinal position, velocity, and measured acceleration of the corresponding vehicle. The relative bearing is depicted by the green line. The red ellipses visualize positions of equal distance value for both vehicles.}
	\label{fig:input_features}
\end{figure}

We retain the core idea of the graph-based scene representation from \cite{klimke2022cooperative}, denoted as $(V,\,E,\,U) \in S$.
Each vehicle in the scene is mapped to a vertex in the set $V$, each of which stores the corresponding input features.
$E$ denotes the set of edges, which we now enhance by a vector of edge features:
\begin{equation}
(v_i,\,v_j,\,g_{ij},\,r) \in E,
\end{equation}
where $v_i$ and $v_j$ denote the source and destination vertex, respectively, while $r \in U = \{\text{same~lane},\,\text{crossing}\}$ specifies the edge type.
The edge feature $g_{ij}$ will be described below.
Two vehicles are connected by an edge in the graph, if and only if their paths require coordination to safely pass the intersection.
In case both vehicles are on different lanes and have a conflict point ahead, the corresponding vertices are bidirectionally connected by \emph{crossing} edges, as shown in Fig.~\ref{fig:input_features}.
If the two vehicles are currently driving on the same path, a \emph{same lane} edge is added to the graph pointing from the predecessor to the following vehicle.

We propose to remove the distance measure used in \cite{klimke2022cooperative} from the vertex input features and replace this information by the novel edges features.
The longitudinal position and scalar velocity are retained and complemented by the current acceleration measurement $\tilde{a}$ for each vehicle, resulting in the vertex input feature vector $\boldsymbol{h}^{(0)} = [s,\,v,\,\tilde{a}]^T$.
The upper index $(0)$ denotes the input layer of the GNN and the tilde is to differentiate the measured acceleration from the action output.
Each edge is assigned a two-element input feature vector $\boldsymbol{g}_{ij}^{(0)} = [1/d_{ij},\,\chi_{ij}]^T$ for an edge pointing from vehicle~$i$ to vehicle~$j$.
In the present work, the distance measure introduced in \cite{klimke2022cooperative}, based on the Mahalanobis distance, is used and denoted as $d_{ij}$.
By specifying the distance on each edge, the network receives more information than a single aggregated distance on each vertex can convey.
Moreover, the relative bearing to the vehicle from which the edge originates is included to improve distinction of different traffic scenes.
The bearing from vehicle~$j$ to vehicle~$i$ is given by
\begin{equation}
\chi_{ij} = \arctan\left(\frac{\boldsymbol{p}_{i,y} - \boldsymbol{p}_{j,y}}{\boldsymbol{p}_{i,x} - \boldsymbol{p}_{j,x}}\right) - \psi_j,
\end{equation}
where $\boldsymbol{p}_{i,x}$ and $\boldsymbol{p}_{i,y}$ denote, respectively, the $x$ and~$y$ components of the 2D position of vehicle $i$.
The heading of vehicle~$j$ in the world frame is specified by $\psi_j$.
Figure~\ref{fig:input_features} illustrates the composition of these edge feature values.
All of the above calculations assume the vehicle reference point to be located at the center of its body on the ground plane.

\subsection{Network Architecture}
\label{ssec:network}

To accommodate the different edge types in the scene graph representation, the GNN is constructed using relational graph convolution (RGCN) layers.
Those layers maintain a separate weight matrix per edge type for constructing the messages exchanged in a forward pass \cite{gangemi2018modeling}.
Moreover, we propose to include the edge features alongside the encoded feature vector of the source node in the first round of message passing.
This integration is similar to the node update proposed in \cite{shi2021masked}, but concatenates the feature vectors instead of taking the sum.
For the target node update, incoming messages are aggregated by taking the element-wise maximum:
\begin{equation}
\label{eq:update}
\boldsymbol{h}_i^{(2)} = \sigma \left( \sum_{r \in U} \max_{j \in \mathcal{N}_i^r} \boldsymbol{W}_{\!\! r}^{(1)} [\boldsymbol{h}_j^{(1)}, \boldsymbol{e}_{ji}^{(1)}] + \boldsymbol{W}_{\! 0}^{(1)} \boldsymbol{h}_i^{(1)} \right),
\end{equation}
where $[\cdot,\cdot]$ denotes the concatenation of two vectors.
All nodes that have an outgoing edge of type~$r$ connected to the target node~$i$ are contained in the set $\mathcal{N}_i^r$.
The weight matrix for each edge type $r \in U$ is called $\boldsymbol{W}_{\!\! r}$, while the previous target node vector is multiplied by $\boldsymbol{W}_{\! 0}$.
We choose the rectified linear unit (ReLU) as the non-linear activation function~$\sigma$ on the basis of empiric observations.

\begin{figure}
	\centering
	\includegraphics[width=\linewidth]{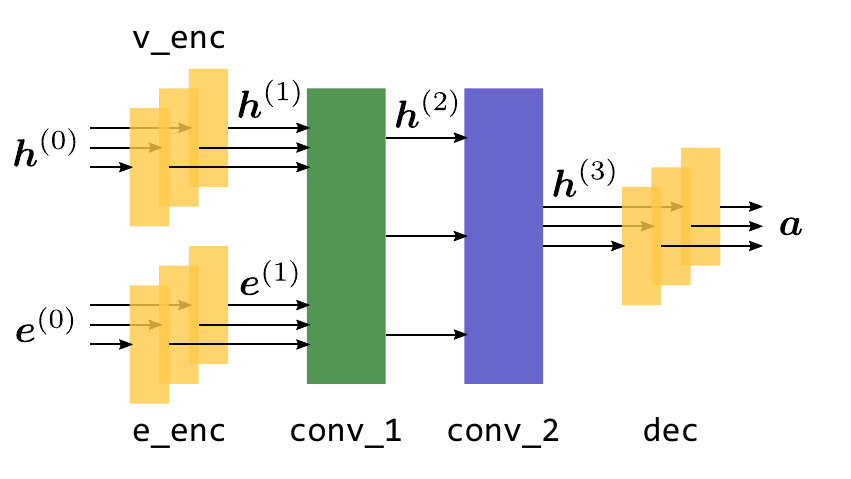}
	\caption{The GNN architecture for the actor network maps input vertex features $\boldsymbol{h}^{(0)}$ and edge features $\boldsymbol{e}^{(0)}$ to one joint action $\boldsymbol{a}$. Fully connected layers are depicted in yellow. The edge feature-enhanced RGCN layer according to \eqref{eq:update} is shown in green and the ordinary RGCN layer in blue.}
	\label{fig:actor}
\end{figure}

It can be observed from \eqref{eq:update} that this layer consumes vertex features and edge features, but only outputs vertex features.
In the present work, this is a reasonable approach, because there is nothing to infer on the edges.
Ultimately, each vertex infers a desired acceleration for the corresponding vehicle, or the vertex features are aggregated to form a joint Q~value estimate.
The overall network architecture is thus analogous to the one presented in \cite{klimke2022cooperative}, but the first graph convolutional layer is replaced by the modified layer to support edge features.
First, the vertex input vectors are encoded by the 64-unit fully connected layer \texttt{v\_enc} independently with shared weights.
Similarly, the edge features are processed by the 32-unit encoder \texttt{e\_enc}.
These intermediate features are then passed to the modified RGCN layer \texttt{conv\_1} alongside the encoded edge features.
Afterwards, only vertex features are preserved that pass a second round of message passing in \texttt{conv\_2} before reaching the action decoder \texttt{dec}, as shown in Fig.~\ref{fig:actor}.

\section{Simulation Environment}
\label{sec:sim}

The training and evaluation environment is based on the open-source Highway-env simulator \cite{highway-env}, which uses the kinematic bicycle model \cite{kong2015kinematic} to compute vehicle motion.
This vehicle model is commonly used for simulating low-dynamics driving.
The simulator was adapted to support centralized multi-agent planning and extended to allow signalized intersection control.
In our environment, the vehicles’ lateral movement is set to follow predetermined paths across the intersection.
The vehicles’ longitudinal motion is determined by the acceleration values provided either by the agents or by a car following (CF) model, which uses its own status (ego) and that of the vehicle in front (leader) as inputs.
To support static priority rules and traffic lights, the CF model was modified to consider right of way.
This means that if a vehicle is yielding, it should come to a halt at the end of the inflow lane before entering the intersection.
To achieve this, the leading vehicle’s values are replaced with the values of a stationary vehicle at the halt point, if the gap between the leader and the ego vehicle is larger than the distance to the halt point.
Therefore, no changes are required to the logic of the CF model.

The CF model used by the static priority rule control scheme in \cite{klimke2022cooperative} and for the new signal-controlled intersections was updated from the Intelligent Driver Model (IDM) \cite{treiber2000congested} to the Extended Intelligent Driver Model (EIDM) \cite{salles2020extending}.
Compared to the IDM, the behavior produced by the EIDM more closely resembles human drivers, which was validated by comparison to aerial traffic observations \cite{salles2020extending}.
To ensure that the improved drive-off behavior presented in the paper occurs in all drive-off situations, the minimum gap calculation of the EIDM was adapted.
In practice, this results in less drive-off delay and thus higher vehicle throughput.
The parameters from \cite{salles2020extending} were retained, except for maximum and minimum acceleration, which were adjusted to be comparable with the RL agent.

\begin{figure*}
	\centering
	\subfloat[Small 3-way (S)]{\includegraphics[width=0.24\textwidth]{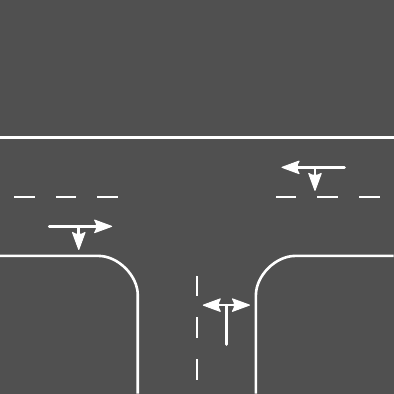}\label{sfig:intersection_s}}
	\hfill
	\subfloat[Medium 4-way (M)]{\includegraphics[width=0.24\textwidth]{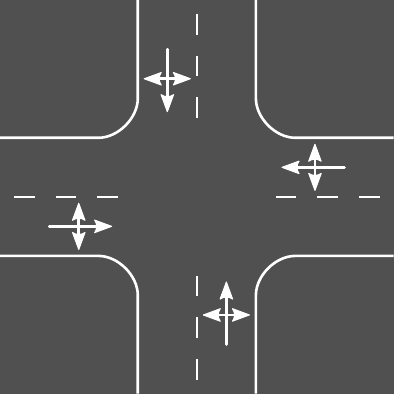}\label{sfig:intersection_m}}
	\hfill
	\subfloat[Large 4-way (L)]{\includegraphics[width=0.24\textwidth]{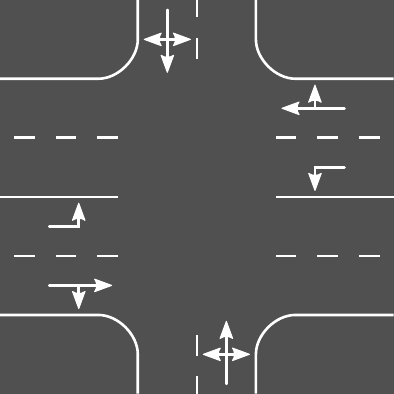}\label{sfig:intersection_l}}
	\hfill
	\subfloat[Extra large 4-way (XL)]{\includegraphics[width=0.24\textwidth]{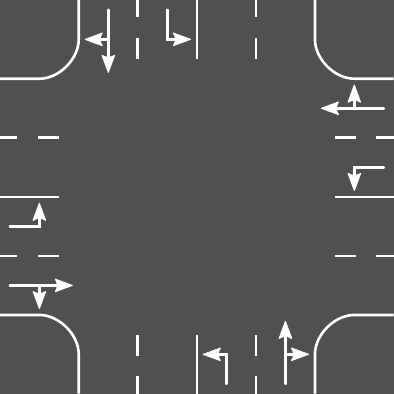}\label{sfig:intersection_xl}}
	\caption{The four intersection layouts used within this study.}
	\label{fig:intersections}
\end{figure*}

Intersections of different sizes are used to simulate varying amounts of traffic demand and complexity.
All intersections feature one horizontal main road and one vertical side road.
When using yielding traffic control, vehicles from the side road must give way to the traffic on the main road.
The layouts of all intersections used within this study are shown in Fig.~\ref{fig:intersections}.
Besides a small 3-way junction (S) and a medium intersection (M) with one lane per driving direction, a large intersection (L) with separate left-turn lanes on the main road as well as an extra-large intersection (XL) with left-turn lanes for both roads are investigated.

\begin{figure}
	\centering
\begingroup%
  \makeatletter%
  \providecommand\color[2][]{%
    \errmessage{(Inkscape) Color is used for the text in Inkscape, but the package 'color.sty' is not loaded}%
    \renewcommand\color[2][]{}%
  }%
  \providecommand\transparent[1]{%
    \errmessage{(Inkscape) Transparency is used (non-zero) for the text in Inkscape, but the package 'transparent.sty' is not loaded}%
    \renewcommand\transparent[1]{}%
  }%
  \providecommand\rotatebox[2]{#2}%
  \newcommand*\fsize{\dimexpr\f@size pt\relax}%
  \newcommand*\lineheight[1]{\fontsize{\fsize}{#1\fsize}\selectfont}%
  \ifx\svgwidth\undefined%
    \setlength{\unitlength}{244.80000173bp}%
    \ifx\svgscale\undefined%
      \relax%
    \else%
      \setlength{\unitlength}{\unitlength * \real{\svgscale}}%
    \fi%
  \else%
    \setlength{\unitlength}{\svgwidth}%
  \fi%
  \global\let\svgwidth\undefined%
  \global\let\svgscale\undefined%
  \makeatother%
  \begin{picture}(1,0.45391385)%
    \lineheight{1}%
    \setlength\tabcolsep{0pt}%
    \put(0,0){\includegraphics[width=\unitlength,page=1]{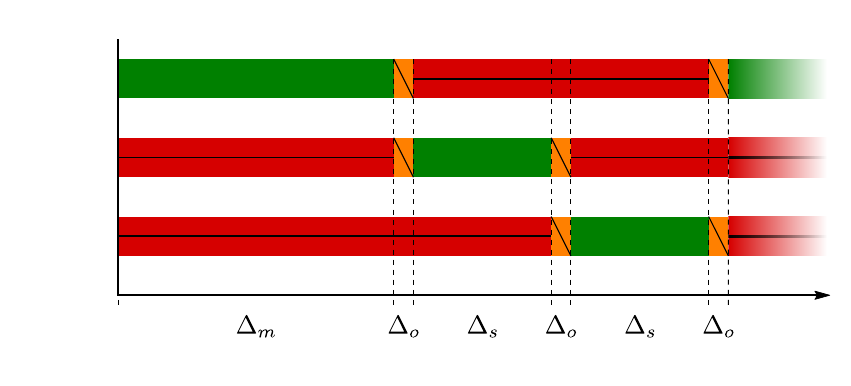}}%
    \put(0.00998590,0.34975789){\color[rgb]{0,0,0}\makebox(0,0)[lt]{\lineheight{1.25}\smash{\begin{tabular}[t]{l}\footnotesize Group 1\end{tabular}}}}%
    \put(0.00881326,0.25710167){\color[rgb]{0,0,0}\makebox(0,0)[lt]{\lineheight{1.25}\smash{\begin{tabular}[t]{l}\footnotesize Group 2\end{tabular}}}}%
    \put(0.00910297,0.16446621){\color[rgb]{0,0,0}\makebox(0,0)[lt]{\lineheight{1.25}\smash{\begin{tabular}[t]{l}\footnotesize Group 3\end{tabular}}}}%
    \put(0.96867150,0.07432103){\color[rgb]{0,0,0}\makebox(0,0)[lt]{\lineheight{1.25}\smash{\begin{tabular}[t]{l}\footnotesize t\end{tabular}}}}%
  \end{picture}%
\endgroup%

	\caption{Traffic light pattern for the large intersection. Group~1 controls the main road outer lanes. Group~2 corresponds to the main road left-turn lanes and Group~3 to the side road.}
	\label{fig:trafficlight}
\end{figure}
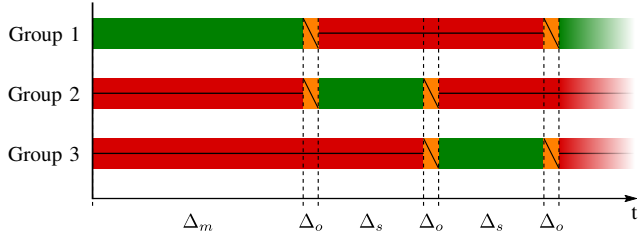

As of today, traffic lights (TL) are the prevalent choice for managing traffic at large urban intersections.
To assess the relative performance of the proposed learned planning policy, our simulation environment was modified to support signalized intersection control.
The traffic in each inflow lane is controlled by a traffic light.
The timings of the traffic lights are constant and depend on the lane's traffic demand.
The main road has green phases of $\Delta_m = \SI{24}{\second}$, while the side road and left-turn lanes have green phases of $\Delta_s = \SI{12}{\second}$.
The yellow phase is $\Delta_o = \SI{2}{\second}$, during which vehicles must come to a stop if they can safely do so.
Vehicles only start accelerating when the signal is green, and an all-red phase is omitted.
The left-turn lanes on the large and extra-large intersections have a separate traffic light with its own green phase.
The configuration used for the large intersection is exemplarily shown in Fig.~\ref{fig:trafficlight}.
Vehicles turning left halt only after entering the intersection, when yielding to oncoming traffic, and clear it at the latest at the end of the green phase.

\section{Evaluation}
\label{sec:eval}

The proposed approach is evaluated using the simulation environment introduced in Section~\ref{sec:sim}.
We present a detailed analysis on the traffic management performance in comparison to multiple baselines in Section~\ref{ssec:baselines}.
Afterwards, Section~\ref{ssec:generalization} demonstrates the generalization capabilities of the graph-based representation to intersection layouts not seen during training.

\subsection{Comparison to Baselines}
\label{ssec:baselines}

To assess the performance of the proposed enhanced RL~planner (eRL), the model was trained and evaluated on the multi-lane four-way intersection layout depicted in Fig.~\subref*{sfig:intersection_l}.
We benchmark against the following baselines:
\begin{itemize}
\item A traditional TL~controller as introduced in Section~\ref{sec:sim},
\item A FIFO policy handling vehicles in order of appearance,
\item An enhanced first-in-first-out (eFIFO) scheme,
\item The legacy RL~planner model from \cite{klimke2022cooperative}.
\end{itemize}
In contrast to a classical FIFO policy, the eFIFO does not enforce a strict ordering on the whole intersection, but only within groups of vehicles that share conflict points.
Technically, the eFIFO considers the vehicles' distance to the intersection for prioritization.
Thereby, a convoy of vehicles on a common lane may traverse the intersection in one go, surpassing any vehicles in other lanes that may be waiting longer.
Accepting this kind of priority inversion for the sake of overall performance gain might lead to increased waiting times for individual vehicles.

\begin{figure}
	\centering
	\includegraphics[width=\linewidth]{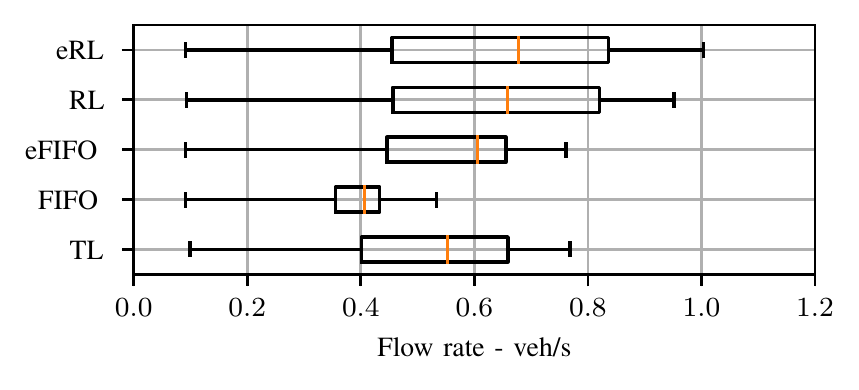}
	\caption{Flow rate distribution for all baselines and learned planner models over varying traffic conditions on the L~intersection.}
	\label{fig:flow_rate}
\end{figure}

In simulation, varying traffic conditions at urban intersections are evaluated by conducting 100 evaluation runs of \SI{100}{\second} length at different vehicle inflow rates.
Hereby, a set of traffic demands between 0.05 and 0.3~vehicles per second and main road lane is sampled.
Like in \cite{klimke2022cooperative}, the rate of traffic entering the simulation is modeled for each lane using an independent shifted exponential distribution, while the vehicle rate is halved for the side roads and left-turn lanes.
All vehicles are spawned \SI{75}{\meter} ahead of the intersection, while the cooperative planner only takes over control \SI{50}{\meter} in front of it.
By skipping vehicles that are farther away, the computational demand is reduced while ensuring that the cooperative planner can influence traffic sufficiently early.
In practice, the control radius of the AIM scheme will be limited by communication and perception range as well as other upstream intersections.
In case a traffic jam forms on an incoming lane up to the spawn point, traffic generation is suspended to prevent immediate collisions.
The \emph{flow rate} describes the number of vehicles that crossed the intersection in a given time frame.
Depending on the used traffic management scheme, different flow rates can be achieved on the same intersection, as depicted in Fig.~\ref{fig:flow_rate}.
The lower bound is virtually identical for all approaches, because in this case, the flow rate is solely determined by the number of incoming vehicles.
It becomes apparent that the plain FIFO falls behind any other approach in terms of vehicle throughput.
Considering the median flow rate, the eFIFO slightly outperforms the TL, which has to rely on its static timing that might not be optimal for each scenario.
The baselines are outperformed by both learned planners, while the eRL~planner proposed in this paper shows a small additional benefit in median and maximum flow rate.

\begin{figure}
	\centering
	\includegraphics[width=\linewidth]{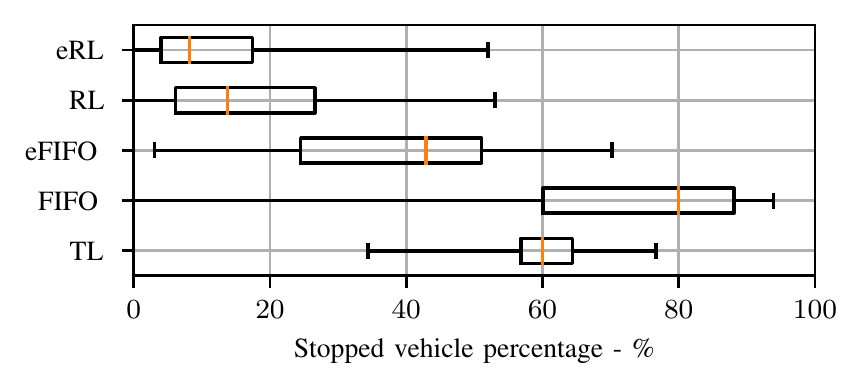}
	\caption{The ratio of vehicles that had to come to a stop while traversing the intersection for all investigated approaches.}
	\label{fig:stop_percentage}
\end{figure}

Apart from the sheer number of vehicles crossing the intersection, ensuring a smooth transition is of interest due to reasons of fuel economy and passenger comfort.
The ideal case is to cross the intersection in uniform motion, which is only possible in practice if there are no vehicles on conflicting paths.
Figure~\ref{fig:stop_percentage} shows a box~plot of the relative numbers of vehicles that had to come to a stop when attempting to cross the intersection.
In this study, a trajectory is considered to contain a stop if the vehicle was moving slower than \SI{0.3}{\meter\per\second} for at least one simulation step.
Both learned planners and the FIFO policy achieve stop-free scenarios, which most likely occur for low traffic density.
With a median of \SI{80}{\percent} stopping trajectories, the FIFO~policy is overall very disruptive to traffic.
Using the eFIFO yields an improvement to a median at roughly \SI{42}{\percent} on a relatively widespread distribution.
The TL~controller inherently causes a large percentage of stopped trajectories, because only vehicles approaching a green light on a free road can cross the intersection without having to stop.
Considering the eRL~planner, in the vast majority of scenarios, less than \SI{20}{\percent} of the vehicles stopped, which is a notable benefit over the legacy model.
In the following, we drop the plain FIFO and legacy RL~planner from detailed analyses, because they are outperformed by their respective enhanced variants.

\begin{figure}
	\centering
	\includegraphics[width=\linewidth]{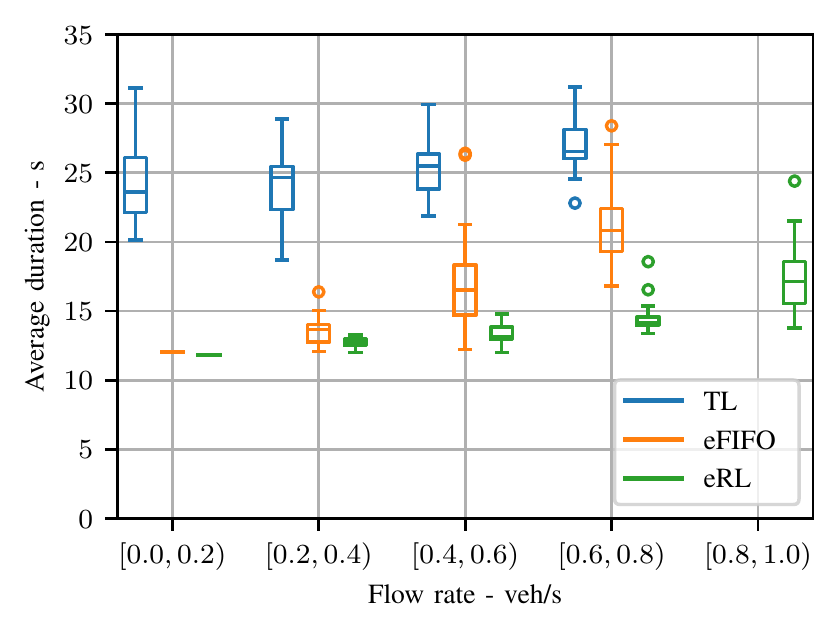}
	\caption{The intersection transition duration experienced at different levels of traffic flow for each intersection management system. A missing box indicates that the corresponding method did not achieve a flow rate in that particular bin.}
	\label{fig:delay_over_flow}
\end{figure}

To assess in more detail how the eRL~planner and the baselines cope with varying traffic densities, we cluster the evaluation results into flow rate bins.
In Fig.~\ref{fig:delay_over_flow}, the duration required by vehicles to cross the intersection is depicted for the TL~controller, the eFIFO scheme and the eRL~planner.
We define \emph{duration} as the time that a vehicle spends in simulation starting at the spawn point until \SI{20}{\meter} beyond the intersection.
It should give an intuition on how disruptive the managed intersection is to traffic flow, with lower values indicating better performance.
The TL~controller achieves a rather constant median duration of about \SI{23}{\second} to \SI{27}{\second}, which corresponds to halve of the signal cycle length (\SI{54}{\second}).
Reducing the cycle length not necessarily leads to lower durations, because the clearance period (yellow light phase) then requires a larger proportion of time, which could ultimately stop all traffic flow.
Both the eFIFO and eRL~planner achieve a minimum around \SI{12}{\second} for low flow rates, which is virtually optimal given the traveled distance.
With rising traffic density, the duration induced by the eFIFO rises rapidly until its capacity limit is reached at around \SI{0.8}{\vehicle\per\second} (cf. also Fig.~\ref{fig:flow_rate}).
Even larger flow rates are only achieved by the eRL~planner that subsequently shows a moderate increase in duration.
Yet, its median duration remains below \SI{18}{\second}, while only few samples exceed \SI{20}{\second}.
This advantage might be explained by the learned behavior to adjust vehicle motion in a way that exploits the available space very efficiently.
Hence, no vehicle has to wait excessively long and a smooth traffic flow is maintained even for high throughput.

\subsection{Generalization Capability}
\label{ssec:generalization}

To assess the generalization capability of the proposed graph-based scene representation, a separate model is trained for each of the intersection layouts shown in Fig.~\ref{fig:intersections}.
Training on the extra large intersection requires an initialization phase on a smaller intersection to converge.
The maximum number of vehicles during training is adapted to the intersection size to keep the traffic from vanishing, while preventing overcrowding, which hinders learning.
Apart from that, the training procedure is identical for each run and matches the description in \cite{klimke2022cooperative}.
For each intersection layout, a set of 100~scenario definitions at a traffic demand in the range $[0.2,\,0.4]$~vehicles per second and major road lane is sampled.
This resembles a dense inflow, which challenges the models' performance at peak load.
Each model is then evaluated on each intersection and scenario configuration, resulting in a total of $16 \times 100$ evaluation runs.

\begin{figure*}
	\centering
	\subfloat[Vehicle collision rate - \%]{\includegraphics[width=0.5\textwidth]{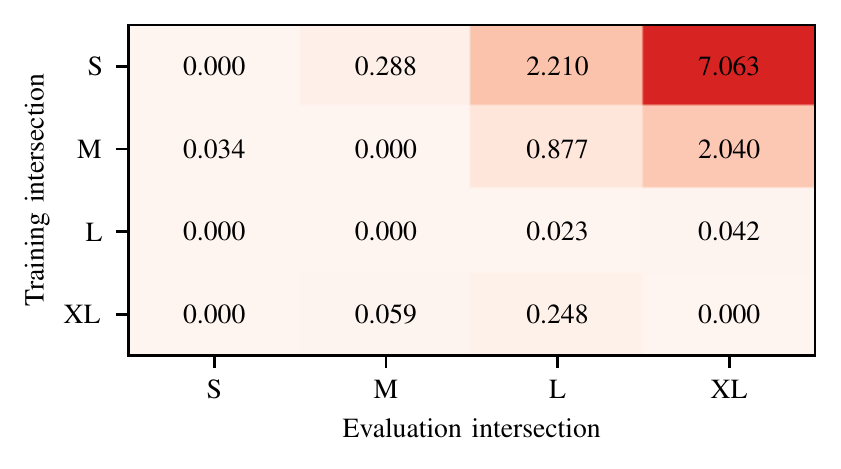}\label{sfig:collisions}}%
	\subfloat[Average flow rate - veh/s]{\includegraphics[width=0.5\textwidth]{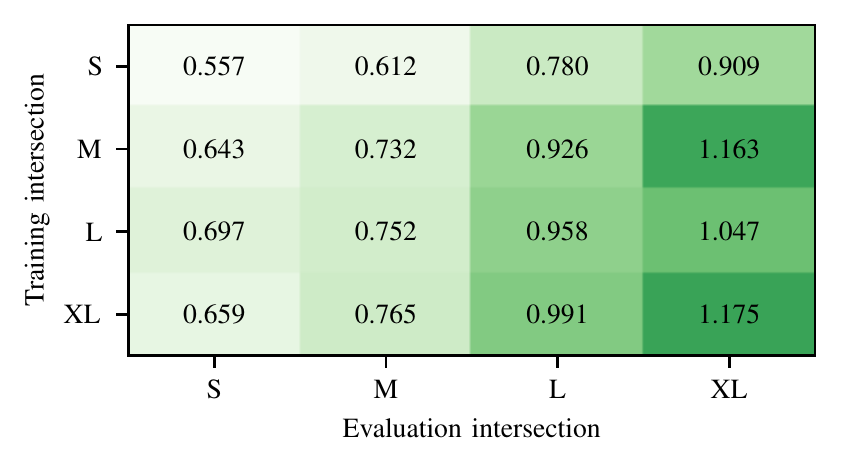}\label{sfig:flowrate}}%
	\caption{Cross validation results for each training intersection and evaluation intersection pair.}
	\label{fig:crossval}
\end{figure*}

The collision rates depicted in Fig.~\subref*{sfig:collisions} indicate, which kind of generalization is feasible.
Models that were trained on an intersection smaller than the one being used for evaluation suffer from an increased number of collisions.
The L and XL~intersections, for instance, feature more complex relations that were not seen during training on the S~intersection.
Especially the additional lane splitting the inflow lane into two, which first appears on the large intersection, seems to pose a hurdle for generalization.
While the M-trained model does not work well on any larger intersection, the L-trained model seems to manage the XL~intersection easily, while being almost collision-free.
Employing a model on an intersection smaller in size than used in training, however, does not lead to increased collision rates as can be observed in the lower triangular matrix.
This might be explained by the fact that all driving relations are retained in the respectively larger intersection layouts.
During training, the model thus had the opportunity to learn a sensible behavior that is leveraged at test time.

Apart from the question to which extend generalization of the proposed models is feasible, the achieved performance is of interest.
For each pair of training intersection and evaluation intersection, Fig.~\subref*{sfig:flowrate} denotes the average flow rate that was observed.
The capacity, i.e., the physically maximum flow rate, of the different intersection layouts rises with increasing size.
Therefore, the highest flow rates were expected to occur during evaluation on the large intersections.
At the same time, models that were trained on comparably smaller intersections typically do not achieve the same level of throughput when evaluated on a larger intersection.
This might be attributed to the lack of geometrical knowledge preventing the efficient usage of additional lanes on the intersection.
A notable exception is the M-trained model used on the XL intersection, which shows an outstanding flow rate.
However, this result should only be considered in conjunction with the induced collision rate, shown in Fig.~\subref*{sfig:collisions}.
It might be possible that the most challenging situations are eliminated by ending in a collision.

\begin{figure}
	\centering
	\includegraphics[width=\linewidth]{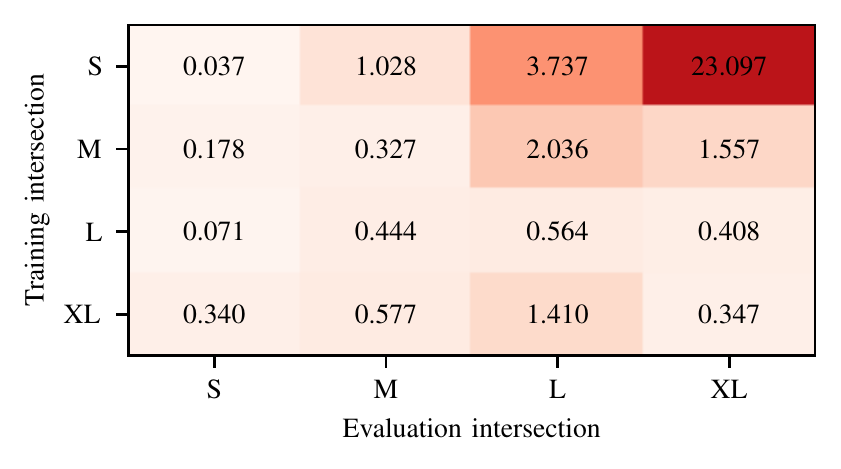}
	\caption{The vehicle collision percentage obtained in cross validation of the legacy model \cite{klimke2022cooperative}.}
	\label{fig:cv_legacy_collision}
\end{figure}

One of the key benefits of the edge-feature-enhanced model is the improved robustness on intersection layouts that were seen during training and those that were not.
This becomes apparent when employing the legacy RL~planner model in the same cross validation experiments.
Figure~\ref{fig:cv_legacy_collision} depicts the collision rates that were obtained under the same evaluation conditions.
Especially the more complex intersection layouts cause an increased collision rate, as can be observed in the right half.
In contrast to the enhanced model, there is no collision-free configuration, not even on the main diagonal corresponding to the models that were trained on the same intersection as used in evaluation.
While the eRL~planner remains mostly collision free in the lower triangular matrix (cf. Fig.~\subref*{sfig:collisions}), the legacy model generalizes not as well to unseen intersection layouts.
The issue becomes even more apparent in the upper triangular matrix.
Notably, all legacy models beside the S-trained one induce less collisions on the XL~intersection than on the L~intersection.
This behavior might be attributed to the stronger symmetry of the XL~intersection leading to less diversity in scenarios, which allows the simpler legacy representation to cope relatively well.

\begin{figure}
	\centering
	\includegraphics[width=\linewidth]{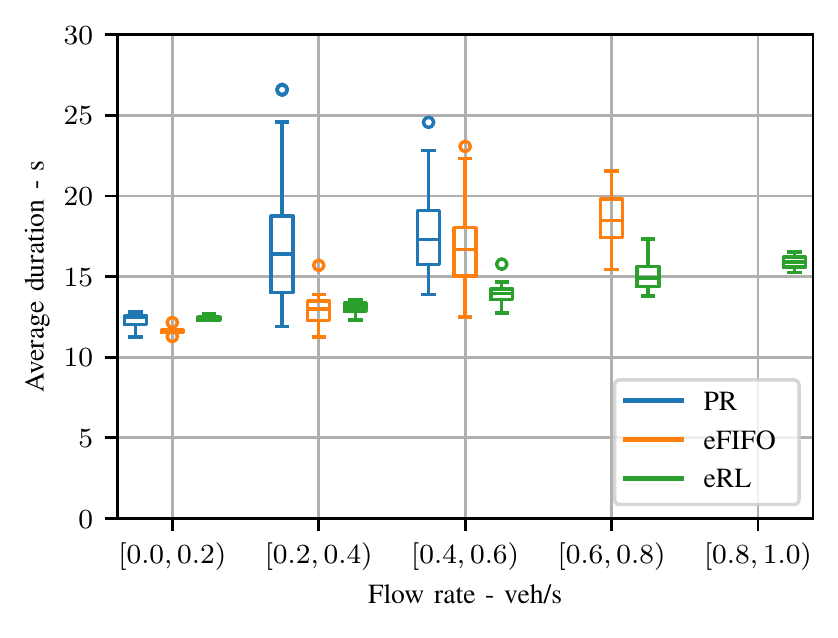}
	\caption{The intersection transition duration experienced at different levels of traffic flow at the medium intersection. The eRL~model used for this experiment was trained on the extra large intersection. Besides the eFIFO, traffic under static priority rules (PR) serves as the second baseline.}
	\label{fig:delay_over_flow_medium}
\end{figure}

Finally, one exemplary generalization of the eRL~planner shall be analyzed in more detail.
We present the induced duration over varying flow rates at the M~intersection when using the model that was trained on the extra large intersection in Fig.~\ref{fig:delay_over_flow_medium}.
The eFIFO serves as the baseline besides traffic governed by static priority rules (PR), which matches the setup in \cite{klimke2022cooperative}.
The PR~baseline was chosen over the TL~controller, because in practice this kind of intersection, under low-density non-automated traffic, is seldom signalized.
For very low traffic densities below \SI{0.2}{\vehicle\per\second}, all approaches show nearly optimal durations.
The learned planner shows a slight deficit compared to the eFIFO, which is presumably due to its anticipation of possible cooperative maneuvers, leading to a more defensive driving style.
Under increasing traffic demand, the PR exhibit a quickly rising duration that can be explained by traffic jams forming on the minor road.
Flow rates above \SI{0.6}{\vehicle\per\second} are unattainable by the PR and lead to a considerable duration increase when using the eFIFO.
The eRL~planner, although not trained on this intersection layout, reaches a vehicle throughput of over \SI{0.8}{\vehicle\per\second} while keeping the duration increase below \SI{5}{\second}.

The RL~planner cannot guarantee collision-free plans, because collision avoidance is learned implicitly via the reward signal.
It practice, this should not be an issue, though, as cooperative plans are subject to sanity checks before being sent out to the vehicles.

\section{Conclusion}
\label{sec:conclusion}

In this work, the GNN-based behavioral planning framework originally presented in \cite{klimke2022cooperative} was improved in key aspects.
By introducing edge features including the pairwise distance measure previously encoded in the vertices, the representation conveys a more descriptive encoding of the scene.
Our proposed model was trained and evaluated on various intersection layouts using the open-source Highway-env simulation environment.
Compared to traffic governed by the TL~controller or the eFIFO scheme, the eRL~planner significantly reduces the duration that the vehicles spend crossing the intersection.
This holds for a broad range of traffic densities, up to flow rate levels that are unattainable by the baselines.
It was shown that the learned model can be deployed to intersection layouts that were not seen during training.
While the generalization towards larger intersections is only feasible to a certain degree, the application to smaller ones is virtually free of restrictions.

In the future, we plan to extend this work to cooperative planning in mixed traffic, i.e. human drivers and automated vehicles sharing the road.
Moreover, the gap between simulation and real-world application is to be shrinked by integrating a dedicated motion planner on detailed vehicle models and a high-fidelity simulator.

\balance
\bibliographystyle{IEEEtran}
\bibliography{references}

\begin{thebibliography}{10}
\providecommand{\url}[1]{#1}
\csname url@samestyle\endcsname
\providecommand{\newblock}{\relax}
\providecommand{\bibinfo}[2]{#2}
\providecommand{\BIBentrySTDinterwordspacing}{\spaceskip=0pt\relax}
\providecommand{\BIBentryALTinterwordstretchfactor}{4}
\providecommand{\BIBentryALTinterwordspacing}{\spaceskip=\fontdimen2\font plus
\BIBentryALTinterwordstretchfactor\fontdimen3\font minus
  \fontdimen4\font\relax}
\providecommand{\BIBforeignlanguage}[2]{{%
\expandafter\ifx\csname l@#1\endcsname\relax
\typeout{** WARNING: IEEEtran.bst: No hyphenation pattern has been}%
\typeout{** loaded for the language `#1'. Using the pattern for}%
\typeout{** the default language instead.}%
\else
\language=\csname l@#1\endcsname
\fi
#2}}
\providecommand{\BIBdecl}{\relax}
\BIBdecl

\bibitem{buchholz2021handling}
M.~Buchholz, J.~C. M\"uller, M.~Herrmann, J.~Strohbeck, B.~V\"olz, M.~Maier,
  J.~Paczia, O.~Stein, H.~Rehborn, and R.-W. Henn, ``Handling {Occlusions} in
  {Automated} {Driving} {Using} a {Multiaccess} {Edge} {Computing}
  {Server}-{Based} {Environment} {Model} {From} {Infrastructure} {Sensors},''
  \emph{IEEE Intelligent Transportation Systems Magazine}, to be published,
  doi: 10.1109/MITS.2021.3089743.

\bibitem{klimke2022cooperative}
\BIBentryALTinterwordspacing
M.~Klimke, B.~V\"olz, and M.~Buchholz, ``Cooperative {Behavioral} {Planning}
  for {Automated} {Driving} using {Graph} {Neural} {Networks},'' in \emph{2022
  {IEEE} {Intelligent} {Vehicles} {Symposium} ({IV})}, to be published.
  [Online]. Available: \url{https://arxiv.org/abs/2202.11376}
\BIBentrySTDinterwordspacing

\bibitem{nichting2019explicit}
M.~Nichting, D.~Hes, J.~Schindler, T.~Hesse, and F.~Koster, ``Explicit
  {Negotiation} {Method} for {Cooperative} {Automated} {Vehicles},'' in
  \emph{2019 {IEEE} {International} {Conference} on {Vehicular} {Electronics}
  and {Safety} ({ICVES})}.\hskip 1em plus 0.5em minus 0.4em\relax IEEE, Sep.
  2019, pp. 1--7.

\bibitem{nichting2020space}
M.~Nichting, D.~Hess, J.~Schindler, T.~Hesse, and F.~Koster, ``Space {Time}
  {Reservation} {Procedure} ({STRP}) for {V2X}-{Based} {Maneuver}
  {Coordination} of {Cooperative} {Automated} {Vehicles} in {Diverse}
  {Conflict} {Scenarios},'' in \emph{2020 {IEEE} {Intelligent} {Vehicles}
  {Symposium} ({IV})}.\hskip 1em plus 0.5em minus 0.4em\relax IEEE, Oct. 2020,
  pp. 502--509.

\bibitem{dresner2008multiagent}
K.~Dresner and P.~Stone, ``A {Multiagent} {Approach} to {Autonomous}
  {Intersection} {Management},'' \emph{Journal of Artificial Intelligence
  Research}, vol.~31, pp. 591--656, Mar. 2008.

\bibitem{li2018near-optimal}
B.~Li, Y.~Zhang, Y.~Zhang, N.~Jia, and Y.~Ge, ``Near-{Optimal} {Online}
  {Motion} {Planning} of {Connected} and {Automated} {Vehicles} at a
  {Signal}-{Free} and {Lane}-{Free} {Intersection},'' in \emph{2018 {IEEE}
  {Intelligent} {Vehicles} {Symposium} ({IV})}.\hskip 1em plus 0.5em minus
  0.4em\relax IEEE, Jun. 2018, pp. 1432--1437.

\bibitem{malikopoulos2018decentralized}
A.~A. Malikopoulos, C.~G. Cassandras, and Y.~J. Zhang,
  ``\BIBforeignlanguage{en}{A {Decentralized} {Energy}-{Optimal} {Control}
  {Framework} for {Connected} {Automated} {Vehicles} at {Signal}-{Free}
  {Intersections}},'' \emph{\BIBforeignlanguage{en}{Automatica}}, vol.~93, pp.
  244--256, Jul. 2018.

\bibitem{kamal2015vehicle-intersection}
M.~A.~S. Kamal, J.-i. Imura, T.~Hayakawa, A.~Ohata, and K.~Aihara, ``A
  {Vehicle}-{Intersection} {Coordination} {Scheme} for {Smooth} {Flows} of
  {Traffic} {Without} {Using} {Traffic} {Lights},'' \emph{IEEE Transactions on
  Intelligent Transportation Systems}, vol.~16, no.~3, pp. 1136--1147, Jun.
  2015.

\bibitem{zhong2020autonomous}
Z.~Zhong, M.~Nejad, and E.~E. Lee, ``Autonomous and {Semi}-{Autonomous}
  {Intersection} {Management}: {A} {Survey},'' \emph{IEEE Intelligent
  Transportation Systems Magazine}, 2020.

\bibitem{morales_medina2018cooperative}
A.~I. Morales~Medina, N.~van~de Wouw, and H.~Nijmeijer, ``Cooperative
  {Intersection} {Control} {Based} on {Virtual} {Platooning},'' \emph{IEEE
  Transactions on Intelligent Transportation Systems}, vol.~19, no.~6, pp.
  1727--1740, Jun. 2018.

\bibitem{gradinescu2007adaptive}
V.~Gradinescu, C.~Gorgorin, R.~Diaconescu, V.~Cristea, and L.~Iftode,
  ``Adaptive {Traffic} {Lights} {Using} {Car}-to-{Car} {Communication},'' in
  \emph{2007 {IEEE} 65th {Vehicular} {Technology} {Conference} -
  {VTC}2007-{Spring}}.\hskip 1em plus 0.5em minus 0.4em\relax IEEE, Apr. 2007,
  pp. 21--25.

\bibitem{zhu2021survey}
Z.~Zhu and H.~Zhao, ``A {Survey} of {Deep} {RL} and {IL} for {Autonomous}
  {Driving} {Policy} {Learning},'' \emph{IEEE {Transactions} on {Intelligent}
  {Transportation} {Systems}}, to be published, doi: 10.1109/TITS.2021.3134702.

\bibitem{capasso2021end--end}
A.~P. Capasso, P.~Maramotti, A.~Dell'Eva, and A.~Broggi, ``End-to-{End}
  {Intersection} {Handling} using {Multi}-{Agent} {Deep} {Reinforcement}
  {Learning},'' in \emph{2021 {IEEE} {Intelligent} {Vehicles} {Symposium}
  {(IV)}}, 2021, pp. 443--450.

\bibitem{hart2020graph}
P.~Hart and A.~Knoll, ``Graph {Neural} {Networks} and {Reinforcement}
  {Learning} for {Behavior} {Generation} in {Semantic} {Environments},'' in
  \emph{2020 {IEEE} {Intelligent} {Vehicles} {Symposium} ({IV})}.\hskip 1em
  plus 0.5em minus 0.4em\relax IEEE, Oct. 2020, pp. 1589--1594.

\bibitem{wu2019dcl-aim}
Y.~Wu, H.~Chen, and F.~Zhu, ``\BIBforeignlanguage{en}{{DCL}-{AIM}:
  {Decentralized} {Coordination} {Learning} of {Autonomous} {Intersection}
  {Management} for {Connected} and {Automated} {Vehicles}},''
  \emph{\BIBforeignlanguage{en}{Transportation Research Part C: Emerging
  Technologies}}, vol. 103, pp. 246--260, Jun. 2019.

\bibitem{fujimoto2018addressing}
S.~Fujimoto, H.~van Hoof, and D.~Meger, ``Addressing {Function} {Approximation}
  {Error} in {Actor}-{Critic} {Methods},'' in \emph{Proceedings of the 35th
  {International} {Conference} on {Machine} {Learning}}, ser. Proceedings of
  {Machine} {Learning} {Research}, J.~Dy and A.~Krause, Eds., vol.~80.\hskip
  1em plus 0.5em minus 0.4em\relax PMLR, Jul. 2018, pp. 1587--1596.

\bibitem{gangemi2018modeling}
M.~Schlichtkrull, T.~N. Kipf, P.~Bloem, R.~van~den Berg, I.~Titov, and
  M.~Welling, ``Modeling {Relational} {Data} with {Graph} {Convolutional}
  {Networks},'' in \emph{The {Semantic} {Web}}, A.~Gangemi, R.~Navigli, M.-E.
  Vidal, P.~Hitzler, R.~Troncy, L.~Hollink, A.~Tordai, and M.~Alam, Eds.\hskip
  1em plus 0.5em minus 0.4em\relax Cham: Springer International Publishing,
  2018, vol. 10843, pp. 593--607.

\bibitem{shi2021masked}
Y.~Shi, Z.~Huang, W.~Wang, H.~Zhong, S.~Feng, and Y.~Sun, ``Masked {Label}
  {Prediction}: {Unified} {Message} {Passing} {Model} for {Semi}-{Supervised}
  {Classification},'' in \emph{Proceedings of the {Thirtieth} {International}
  {Joint} {Conference} on {Artificial} {Intelligence}, {IJCAI-21}}, Z.-H. Zhou,
  Ed., Aug. 2021, pp. 1548--1554.

\bibitem{highway-env}
\BIBentryALTinterwordspacing
E.~Leurent, ``An {Environment} for {Autonomous} {Driving}
  {Decision}-{Making},'' May 2018. [Online]. Available:
  \url{https://github.com/eleurent/highway-env}
\BIBentrySTDinterwordspacing

\bibitem{kong2015kinematic}
J.~Kong, M.~Pfeiffer, G.~Schildbach, and F.~Borrelli, ``Kinematic and {Dynamic}
  {Vehicle} {Models} for {Autonomous} {Driving} {Control} {Design},'' in
  \emph{2015 {IEEE} {Intelligent} {Vehicles} {Symposium} ({IV})}.\hskip 1em
  plus 0.5em minus 0.4em\relax IEEE, Jun. 2015, pp. 1094--1099.

\bibitem{treiber2000congested}
M.~Treiber, A.~Hennecke, and D.~Helbing, ``\BIBforeignlanguage{en}{Congested
  {Traffic} {States} in {Empirical} {Observations} and {Microscopic}
  {Simulations}},'' \emph{\BIBforeignlanguage{en}{Physical Review E}}, vol.~62,
  no.~2, pp. 1805--1824, Aug. 2000.

\bibitem{salles2020extending}
D.~Salles, S.~Kaufmann, and H.-C. Reuss, ``Extending the {Intelligent} {Driver}
  {Model} in {SUMO} and {Verifying} the {Drive} {Off} {Trajectories} with
  {Aerial} {Measurements},'' in \emph{{SUMO} {User} {Conference}}, 2020.

\end{thebibliography}

\end{document}